\pdfoutput=1

\documentclass[11pt]{article}

\usepackage[hyphens]{url}
\usepackage[]{EMNLP2023}

\usepackage{times}
\usepackage{latexsym}

\usepackage[T1]{fontenc}

\usepackage[utf8]{inputenc}

\usepackage{microtype}

\usepackage{inconsolata}

\usepackage{xspace}
\usepackage{graphicx}
\usepackage{caption}
\usepackage{multirow}
\usepackage{multicol}
\usepackage{makecell}
\usepackage{booktabs}
\usepackage{tabularx}
\usepackage{adjustbox}
\usepackage{pifont}
\usepackage{soul}
\usepackage{enumitem}
\usepackage{amsmath}
\usepackage{amssymb}
\usepackage{comment}
\usepackage{enumitem}

\newcolumntype{b}{X}
\newcolumntype{s}{>{\hsize=.5\hsize}X}

\definecolor{violet-5}{RGB}{132, 94, 247}

\newcommand{\framework}{\textsc{CO\textsubscript{3}}\xspace}
\newcommand{\dataset}{\textsc{Soda}\xspace}
\newcommand{\model}{\textsc{Cosmo}\xspace}
\newcommand{\modelxxl}{\textsc{Cosmo-11B}\xspace}
\newcommand{\modelxl}{\textsc{Cosmo-3B}\xspace}
\newcommand{\atomictenx}{Atomic\textsuperscript{\textbf{10x}}\xspace}

\newcommand{\frameworkEmoji}{\includegraphics[height=.9em,trim=0 .4em 0 0]{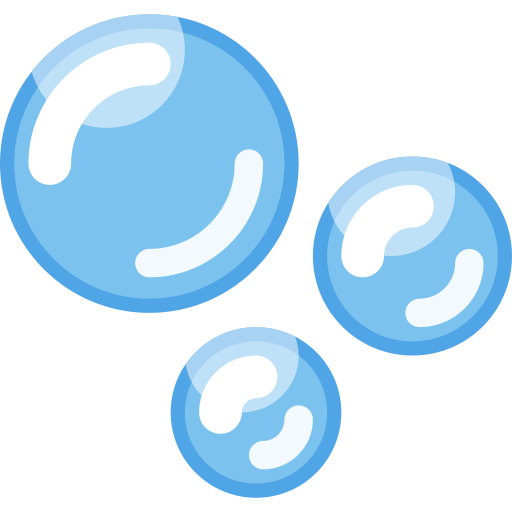}}
\newcommand{\dataEmoji}{\includegraphics[height=.9em,trim=0 .4em 0 0]{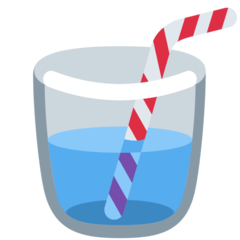}}
\newcommand{\symbolicEmoji}{\includegraphics[height=1.1em,trim=0 .4em 0 0]{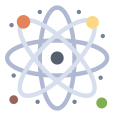}}
\newcommand{\narrativeEmoji}{\includegraphics[height=1.1em,trim=0 .4em 0 0]{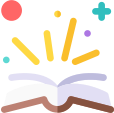}}
\newcommand{\dialogueEmoji}{\includegraphics[height=1.1em,trim=0 .4em 0 0]{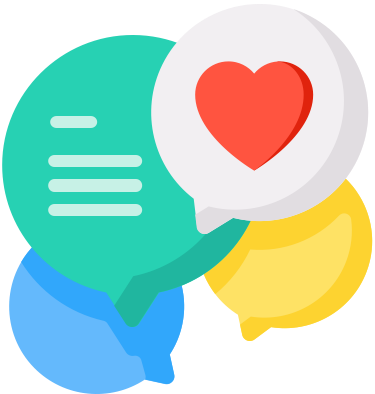}}
\newcommand{\modelEmoji}{\includegraphics[height=1.1em,trim=0 .4em 0 0]{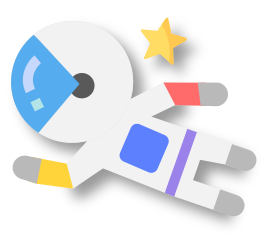}}

\newcommand{\frameworkWithEmoji}{\frameworkEmoji\xspace\textsc{CO\textsubscript{3}}\xspace}
\newcommand{\datasetWithEmoji}{\dataEmoji\xspace\textsc{Soda}\xspace}
\newcommand{\modelWithEmoji}{\modelEmoji\xspace\textsc{Cosmo}\xspace}
\newcommand{\eg}{e.g.,\xspace}
\newcommand{\ie}{i.e.,\xspace}

\newcommand{\tabitem}{~~\llap{\textbullet}~~}

\title{\datasetWithEmoji: Million-scale Dialogue Distillation \\ with Social Commonsense Contextualization}

\author{
Hyunwoo Kim$^{\heartsuit \spadesuit}$ \quad
Jack Hessel$^{\heartsuit}$ \quad
Liwei Jiang$^{\heartsuit\diamondsuit}$ \quad
Peter West$^{\diamondsuit}$ \quad
Ximing Lu$^{\diamondsuit}$ \\
\textbf{Youngjae Yu}$^{\heartsuit}$ \quad
\textbf{Pei Zhou}$^{\heartsuit \clubsuit}$ \quad
\textbf{Ronan Le Bras}$^{\heartsuit}$ \quad
\textbf{Malihe Alikhani}$^{\dagger}$ \\
\textbf{Gunhee Kim}$^{\spadesuit}$ \quad
\textbf{Maarten Sap}$^{\heartsuit\ddagger}$ \quad
\textbf{Yejin Choi}$^{\heartsuit\diamondsuit}$ 
\\
\small{$\heartsuit$ Allen Institute for Artificial Intelligence} \quad
\small{$\spadesuit$ Seoul National University} \quad
\small{$\diamondsuit$ University of Washington}\\
\small{$\clubsuit$ University of Southern California} \quad
\small{$\dagger$ University of Pittsburgh} \quad
\small{$\ddagger$ Carnegie Mellon University}
}

\begin{document}
\maketitle
\begin{abstract}

Data scarcity has been a long standing issue in the field of open-domain social dialogue.
To quench this thirst, we present \datasetWithEmoji: the first publicly available, million-scale high-quality social dialogue dataset. 
By contextualizing social commonsense knowledge from a knowledge graph, we are able to distill an exceptionally broad spectrum of social interactions from a large language model.
Human evaluation shows that conversations in \dataset are more consistent, specific, and (surprisingly) \emph{natural} than those in prior human-authored datasets. 

Using \dataset, we train \modelWithEmoji: a generalizable conversation model that is significantly more natural and consistent on unseen datasets than best-performing conversation models (\eg GODEL, BlenderBot-1, Koala, Vicuna).
Experiments reveal \model is sometimes even preferred to the original human-written gold responses.
Additionally, our results shed light on the distinction between knowledge-enriched conversations and natural social chitchats.
We make our data, models, and code public.\footnote{\url{https://hyunw.kim/sodaverse}}

\end{abstract}

\section{Introduction}

Conversations that occur in everyday spoken situations are often not recorded as data. And when they are, such as in the case of text messages, research use is rightly  restricted due to privacy and legal concerns.
As a result, collecting high-quality, everyday social conversations on a large scale has long been recognized as a difficult task \cite{smith-etal-2020-put}.
Previous studies have relied on crowdsourcing focused on specific themes of dialogue (\eg persona, empathy; \citealp{zhang-etal-2018-personalizing, rashkin-etal-2019-towards}).
However, this approach is limited in scale due to its associated costs.
As a result, the progress made in machine dialogues, including generation, evaluation, and understanding, has been severely hindered by the reliance on these small datasets \citep{kann-etal-2022-open, mehri2022report}.

\begin{figure}[t!] \begin{center}
    \includegraphics[width=\columnwidth]{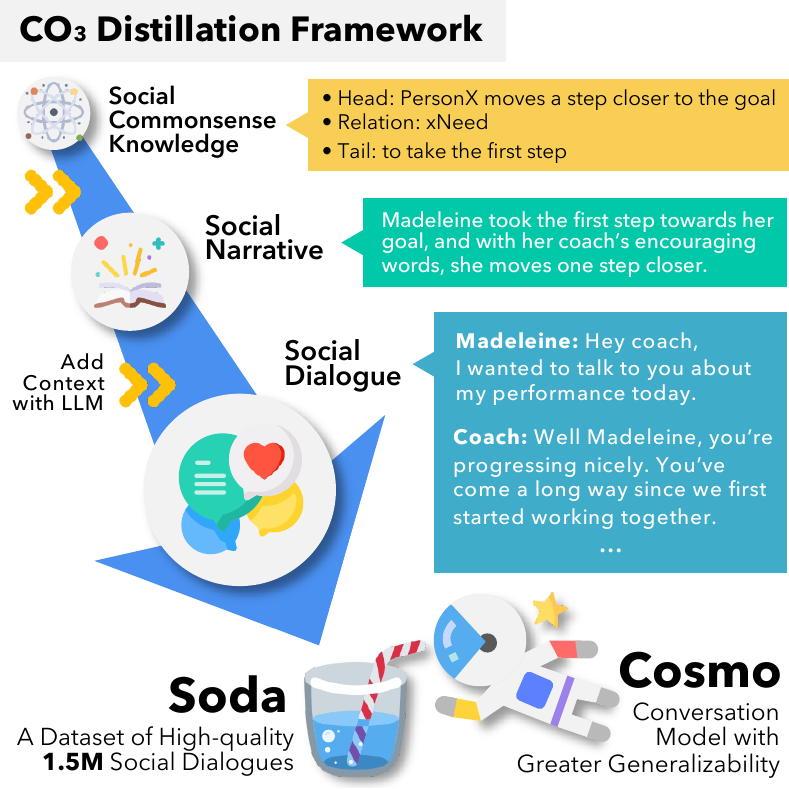}
    \caption{
        An illustration of our \framework framework (\S \ref{sec:contextualization}), \dataset dataset (\S \ref{sec:dataset}), and conversation model \model (\S \ref{sec:model}) trained on \dataset.
        Conversations are distilled from a large language model (LLM) by contextualizing social commonsense.
        The full example is in Table \ref{tab:example}.
        }
    \vspace{-9.5pt}
    \label{fig:figure1}
\end{center} \end{figure}

To alleviate this bottleneck, we introduce \datasetWithEmoji (\textbf{SO}cial \textbf{D}i\textbf{A}logues), a million-scale %
English dialogue dataset covering a wide variety of social interactions. %
As a result of being grounded on rich social commonsense and narratives, \dataset goes beyond specific skill-focused dialogues and features more general conversations.
Our dataset includes 1.5 million dialogues
distilled from a large language model \citep[in our case, GPT-3.5;][]{ouyang2022training} 
resulting in more than 11 million utterances with 300 million tokens:
\dataset is the largest %
publicly available open-domain social conversation dataset.
Human evaluation shows that \dataset surpasses existing human-authored dialogue corpora %
across axes like consistency, specificity, and (surprisingly, even) naturalness (\S \ref{subsec:vs_humandialogue}).

To make \dataset, we propose \frameworkWithEmoji, a framework for \textbf{CO}ntextualizing \textbf{CO}mmonsense for distilling \textbf{CO}nversations from a large language model (LLM).
Illustrated in Figure \ref{fig:figure1}, \framework infuses commonsense knowledge into dialogues by transforming knowledge triples into narratives, and then into dialogues.
Such an approach offers two significant advantages:
(1) maximizing diversity and (2) minimizing nonsensical conversations.
Although generating content using LLMs is relatively easy, determining how to cover diverse content poses a non-trivial challenge.
We find that sampling from an LLM without contexts results in dull conversations (\S \ref{subsec:why_contextualize}).
Because commonsense knowledge graphs cover a wide range of everyday situations \citep{west-etal-2022-symbolic}, conditioning on them results in a broad spectrum of conversations.
Moreover, since LLMs are prone to hallucinations \citep{weidinger2021ethical}, the seed commonsense knowledge can help them stay on a sensible generation path.

With \dataset, we train a \textbf{CO}nver\textbf{S}ation \textbf{MO}del, \modelWithEmoji. 
Human evaluation results demonstrate that: (1) \model generalizes better to unseen conversations than existing best-performing dialogue models, winning by more than 40\% on average in head-to-head comparisons versus BlenderBot \citep{roller-etal-2021-recipes}, Koala \cite{koala_blogpost_2023}, and Vicuna \cite{vicuna2023}  (\S \ref{subsec:out-domain});
(2) \model outperforms BlenderBot (with the same number of parameters) \emph{on the dataset BlenderBot was trained on,} despite never seeing the corpus (\S \ref{subsec:one-sided-out-domain}); and 
(3) \model responses are even preferred over human-authored, ground-truth responses in DailyDialog \cite{li-etal-2017-dailydialog}, a dataset on which \model was not trained on (\S \ref{subsec:out-domain}).

Finally, the distilled dialogues in \dataset represent a significant resource contribution for open-domain dialogue research.
Most of all, \dataset enables the research community to train smaller dialogue agents with competitive capabilities.
Also, \dataset can help enhance the generalizability of other advancements in the dialogue field (\eg understanding and evaluation), which have relied on existing small datasets.
Lastly, \dataset highlights a dimension where recent LLM-based conversational agents (\eg Koala, Vicuna, and ChatGPT) struggle -- \ie the naturalness of the responses (\S \ref{subsec:out-domain} and \S \ref{subsec:in-domain}).
As these models are designed to provide knowledge-based responses, they may generate responses that are informative but lack the naturalness found in social chitchat.
We plan to publicly release \dataset, \model, and \framework under the permissive license CC-BY-4.0, aiming to address the data scarcity issue in open-domain dialogue.

\section{\frameworkWithEmoji: A Contextualization Framework for Conversation Distillation using Commonsense}
\label{sec:contextualization}

We propose CO\textsubscript{3}, a framework for distilling \textbf{co}nversations from large language models (LLMs) by \textbf{co}ntextualizing (\ie adding more context information) \textbf{co}mmonsense knowledge.
Our goal is to obtain natural conversations covering a wide variety of social interactions. \framework consists of three steps:
(1) Retrieving social commonsense from a symbolic commonsense knowledge graph (\S \ref{subsec:kg}),
(2) converting it into sentence form and generating a narrative from the sentence (\S \ref{subsec:commonsense_to_narrative}), and
(3) inferring the conversation participants from the narrative and derive a conversation grounded in the narrative (\S \ref{subsec:narrative_to_conversation}).
We use GPT-3.5 \citep[\ie text-davinci-002\footnote{\url{https://beta.openai.com/docs/model-index-for-researchers/models-referred-to-as-gpt-3-5}};][]{ouyang2022training} to implement CO\textsubscript{3}, though in practice, a different model could be used.
We use \framework to create \dataset: an example is in Table \ref{tab:example}.
More details can be found in Appendix \ref{app:contextualization}.

\subsection{Inspiration Behind \framework}
\textit{What is at the heart of conversation?}
At its core, a conversation is a fundamental form of social interaction \citep{myllyniemi1986conversation}.
These experiences are abstracted into narratives or scripts \citep{mar2008function, rumelhart1975notes, schank1975scripts}.
Eventually, social experiences form our knowledge for explaining everyday events and inferring the mental states of others \citep{heider1958psychology}.
This inference is coined \textit{attribution} in social psychology \citep{baumeister2017social}, and has been studied in NLP as \textit{social commonsense} \citep{rashkin-etal-2018-modeling, sap-etal-2019-social}.
Inspired by cognitive science, we reverse the abstraction process, starting from social commonsense knowledge in symbolic forms, and unfold rich narratives and conversations that could have initially encapsulated those commonsense knowledge.

\subsection{Commonsense Knowledge Graph}
\label{subsec:kg}
Concretely, we start with a commonsense knowledge graph, which captures various relations of everyday events and inferences on others' mental states in symbolic forms \citep{sap-etal-2019-social, hwang2021comet}.
The knowledge graph is represented by symbolic triples describing two events, denoted as the head and tail, and the relation between those two events, \eg Head: \texttt{PersonX moves a step closer to the goal}, Relation: \texttt{xNeed}, Tail: \texttt{to take the first step}.
We use \atomictenx \citep{west-etal-2022-symbolic} as our knowledge graph:
it includes diverse social (\eg intention, desire, reaction) and event-centered (\eg order of events) commonsense.
Since we are interested in distilling social interactions, we only retrieve triples related to \emph{social} (rather than, e.g., physical) commonsense.\footnote{
We leave relations for physical and event-centered commonsense to potential future work.}

\subsection{Commonsense Knowledge $\rightarrow$ Narrative}
\label{subsec:commonsense_to_narrative}

\paragraph{Triple Form to Sentence Form}
Since commonsense knowledge graphs are represented in symbolic form (\ie triples), we first convert them into simple sentences with templates for each relation.
For example, the commonsense knowledge in Table \ref{tab:example} is converted to ``\textit{Madeleine took the first step. Madeleine moves a step closer to the goal}.''
To make the sentences sound more natural, we replace the person variables (\eg PersonX, PersonY) with Top-1K common names of US SSN applicants ranging from 1990 to 2021.\footnote{\href{https://catalog.data.gov/dataset/baby-names-from-social-security-card-applications-national-data}{\nolinkurl{catalog.data.gov/dataset/baby-names-from-social-security-card-applications-national-data}}} 

\paragraph{Sentence Form to Narrative}

Next, we prompt GPT-3.5 to %
instantiate the sentence-form into a two- or three-sentence short narrative, e.g., for the commonsense example above 
``\textit{
Madeleine took the first step towards her goal, and with her coach’s encouraging words, she moves one step closer.
}'' Prior work has shown that LLMs can effectively generate stories with plausible details that go beyond the contents of the prompt \citep{radford2019gpt2}.

\subsection{Narrative $\rightarrow$ Conversation}
\label{subsec:narrative_to_conversation}

\paragraph{Inferring Conversation Participants}
Inferring the conversation participants from the narrative is straightforward in cases where triples contain two person variables (\ie PersonX and PersonY). %
But for triples that include only one person (e.g., the example in Table \ref{tab:example}), we query GPT-3.5 to predict the other interlocutor (\eg \textit{mom}, \textit{coworker}).

\paragraph{Generating Conversation grounded in Narrative}
With the narrative and speakers as input, we prompt GPT-3.5 to generate a full, multi-turn conversation between the speakers in the context of the narrative. %
We append the first speaker as an utterance prefix to the prompt.
Indicating the speakers with prefixes helps GPT-3.5 generate fluent conversations that alternate between the two.

\section{\datasetWithEmoji: \newline A Million-scale Social Dialogue Dataset}
\label{sec:dataset}

{\renewcommand{\arraystretch}{1}
    \begin{table}[t] \begin{center}
    \small
    \setlength{\tabcolsep}{3pt}
    \begin{tabularx}{\linewidth}{X}
        \toprule
        \symbolicEmoji \xspace\xspace \textbf{Symbolic Social Commonsense Knowledge}: \\
        \tabitem Head: PersonX moves a step closer to the goal  \\
        \tabitem Relation: xNeed  \\
        \tabitem Tail: to take the first step \\
        \textbf{Sentence-form Commonsense}: Madeleine took the first step. Madeleine moves a step closer to the goal.  \\
        \midrule
        \narrativeEmoji \xspace\xspace \textbf{Narrative}:\\ Madeleine took the first step towards her goal, and with her coach's encouraging words, she moves one step closer. \\
        \textbf{Speakers}: Madeleine, Coach \\
        \midrule
        \dialogueEmoji \xspace\xspace \textbf{Conversation}:\\
        \textbf{Madeleine:} Hey coach, I wanted to talk to you about my performance today. I was really pushing myself and I think I did pretty well. But I'm still not quite where I want to be. \\
        \textbf{Coach:} Well Madeleine, you're progressing nicely. You've come a long way since we first started working together. But if you want to reach your full potential, there's still some work to be done. \\
        \textbf{Madeleine:} I know that. And I'm willing to put in the work. It's just that sometimes I feel like I'm not making as much progress as I should be. Maybe I'm not training hard enough? Or maybe my technique is off? \\
        \textbf{Coach:} It could be a number of things, Madeleine. But don't worry, we'll figure it out together. Let's just keep working hard and see how things go. \\
        \textbf{Madeleine:} Alright, coach. Thanks for the talk. \\
        \textbf{Coach:} No problem. See you at practice tomorrow. \\
        \bottomrule
    \end{tabularx}
    \vspace{-5pt}
    \caption{
        A sample from \datasetWithEmoji. 
        More examples can be found in Appendix \ref{app:dataset}.
    }
    \vspace{-15pt}
    \label{tab:example}
\end{center}\end{table}}

We obtain \dataset (\textbf{SO}cial \textbf{D}i\textbf{A}logues), a large-scale high-quality conversation dataset covering a wide range of social interactions, by applying a series of post-processing (\S \ref{subsec:postprocessing}) to the conversations generated from our contextualization framework (\S \ref{sec:contextualization}).
We compare \dataset with existing human-curated dialogue corpora (\S \ref{subsec:vs_humandialogue}) and analyze the effectiveness of contextualization (\S \ref{subsec:why_contextualize}).
Table \ref{tab:example} shows a sample from our dataset.
More details are in Appendix \ref{app:dataset}.

\subsection{Post-processing the Conversations}
\label{subsec:postprocessing}

\paragraph{Basic Filtering}

Starting with an initial set of 2.2 million conversations sampled from GPT-3.5,
we: (1) use lexical pattern matching to filter out conversations with erroneous patterns -- \eg repetition and omission of speaker prefixes (6.3\%);
(2) remove conversations that have less than four turns or more than twenty turns (5.7\%);
(3) remove conversations with more than two speakers (11.3\%);\footnote{Although our pipeline naturally generates multi-party conversations as well, we focus on dyadic dialogues in this work.}
and (4) remove conversations where at least one of the speakers was identified as non-human (\eg broomstick, imaginary friend, dog; 5.6\%).

\begin{figure*}[t!] \begin{center}
    \includegraphics[width=0.95\linewidth]{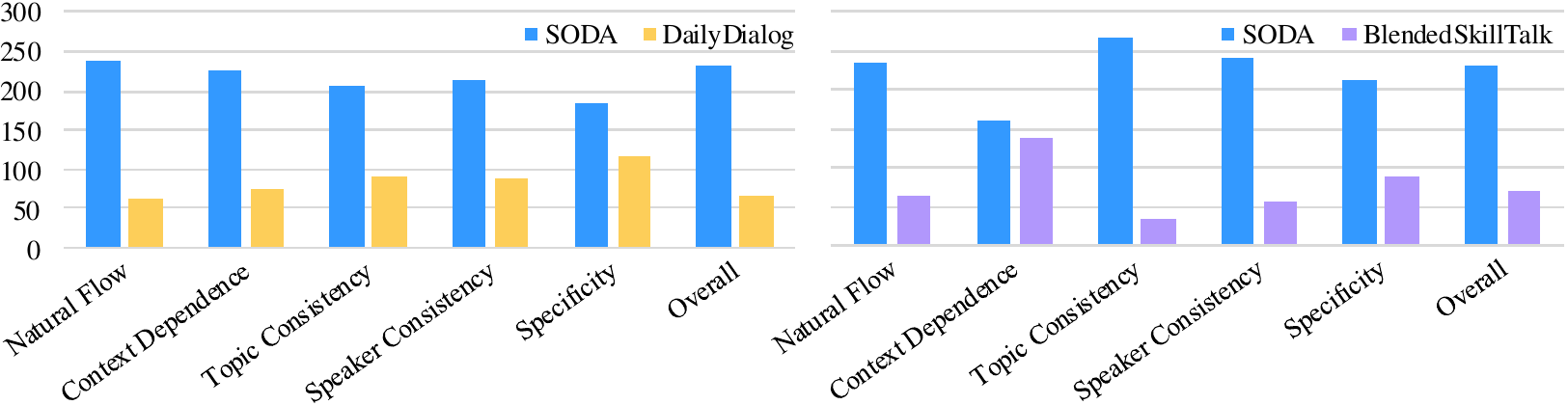}
    \caption{
        Results of head-to-head comparison between dialogues from \datasetWithEmoji, DailyDialog \citep{li-etal-2017-dailydialog}, and BlendedSkillTalk \citep{smith-etal-2020-put} via human judgments (\S \ref{subsec:vs_humandialogue}). 
        The y-axis represents the number of samples preferred by human judges.
        The differences in all of the categories except for the \textit{Context Dependence} comparing \datasetWithEmoji and BlendedSkillTalk are statistically significant ($|z|>3.3, \quad p<0.05$).
        }
    \vspace{-7pt}
    \label{fig:dialogue_quality}
\end{center} \end{figure*}

\paragraph{Safety Filtering}
In order to avoid conversations with dangerous and harmful contents, we apply two safety filters: Canary \citep{kim-etal-2022-prosocialdialog} and Rewire API.\footnote{\url{https://rewire.online/}}
Canary is a narrative dialogue safety model that can classify whether the given context needs caution or intervention.
We discard all conversations marked as needing intervention (usually critical situations, \eg crimes, emergencies; 4.3\%); %
Rewire API is a web-based API for detecting toxic content.
We discard all conversations that are above the threshold of 0.5 for any of the `violence', `hate', and `sexually explicit' criteria ($\sim$1\%).

\paragraph{Commonsense Filtering}
We conduct a small-scale human evaluation via Amazon Mechanical Turk with 100 randomly sampled narrative-conversation pairs (3 annotators per instance) to check whether or not the seed commonsense triple is meaningfully instantiated by the narrative and conversation.
According to majority vote, 88\% of the instances include the seed commonsense knowledge.
Given that the majority of human-annotated samples include the seed commonsense, we focus our filtering on excluding narrative-conversation pairs that lack the head event, as they are irrelevant to the given seed commonsense.

To apply this filter to all entries of the corpus, we use GPT-3.5 as a zero-shot classifier.
As GPT-3.5 demonstrated great performance in question answering \citep{ouyang2022training}, we validate the generated narrative-conversation pairs by asking the language model itself to judge whether or not the head of the commonsense triple is implied.
We formulate this as three-way multiple choice questions (\ie \textit{yes}, \textit{no}, and \textit{unknown}) and rank the answers according to their perplexity scores from GPT-3.5.
This zero-shot classifier achieves high performance on the human-annotated subset, with a precision of 97 for answering ``yes".
We find 95\% of the filtered conversations are identified by GPT-3.5 as containing the head event.
Pairs that lack the head event are removed to ensure relevance between the narrative-conversation pairs and commonsense triples.
More details are in Appendix \ref{app:postprocessing}.

\paragraph{Final Dataset} After all filtering, 68.9\% of the initial conversations remain, which form the 1,486,896 conversations in \dataset.

\paragraph{Name Bias Mitigation}
We aim to minimize biases associated with specific names while increasing inclusion and diversity.
Both language models and curated datasets often exhibit demographic imbalances \citep{dinan-etal-2020-queens, weidinger2021ethical, sheng2021revealing}.
Inspired by \citet{smith2021hi}, we randomly replace all names in conversations with Top-10K names of US SSN applicants from 1990 to 2021.\footnote{We use Top-1K names when contextualizing the commonsense triples in \S \ref{subsec:commonsense_to_narrative}.} 
This covers 95\% of all applicants' names from the chosen time range window, including various names from diverse gender\footnote{Gender-neutral and nonbinary names are also included.}
and ethnic backgrounds.

\subsection{Comparing \dataset with \newline Human-authored Dialogues}
\label{subsec:vs_humandialogue}

\paragraph{High Quality}
To assess relative quality of the corpus, we conduct head-to-head human evaluations on Amazon Mechanical Turk, comparing \dataset with two widely used open-domain dialogue datasets: DailyDialog \citep{li-etal-2017-dailydialog} and BlendedSkillTalk \citep{smith-etal-2020-put}.
We random sample 300 dialogues from each dataset and evaluate them according to six criteria \citep{mehri2022report}: (1) natural flow, (2) context dependence, (3) topic consistency, (4) speaker consistency, (5) specificity, and (6) overall.
Judges are asked to select a better dialogue between the two, regarding each criterion.
For context dependence, we ask the judges to choose which conversation includes responses that are more dependent on previous turns.
Further details are in Appendix \ref{app:vs_humandialogue}.

Despite being fully machine-generated, human raters judge \dataset as better in quality compared to both DailyDialog and BlendedSkillTalk across all axes by a large margin, except for the context dependence comparing with BlendedSkillTalk (see
Figure \ref{fig:dialogue_quality}). %
In particular, evaluators rate the flow of \dataset to be significantly more natural than other human-authored artificial conversation datasets.\footnote{A power analysis suggests that with our setup, 
we can detect effect sizes as small as 0.17 with a power and significance level of 95\% \cite{faul2014g}.}

\paragraph{Large Scale}
With 1.5 million conversations, \dataset is the largest in scale compared to existing crowdsourced open-domain dialogue datasets and the machine-human generated ProsocialDialog dataset (Table \ref{tab:dataset_stats}). 
It contains more than 11 million utterances and each conversation is grounded in a short narrative describing the context.
In total, \dataset consists of 300 million tokens, making it a rich source for training conversation models.

{\renewcommand{\arraystretch}{1.2}
    \begin{table}[t!] \begin{center}
    \begin{adjustbox}{width=\columnwidth}
    \begin{tabular}{lcccc}
        \toprule
                                & \makecell{\#Dialog}         & \makecell{Avg.\\\#Turns}    & \makecell{Avg. Utt.\\Length}  & \makecell{Lexical\\Diversity}   \\ %
        \midrule  %
        \makecell[l]{DailyDialog}             & 13K            & 7.9                       & 14.6                        & 63.0                 \\ %
        \makecell[l]{PersonaChat}             & 11K            & 14.8                      & 14.2                        & 43.6                 \\ %
        \makecell[l]{WizardOfWikipedia}     & 22K            & 9.1                       & 16.4                        & 60.3                   \\ %
        \makecell[l]{EmpatheticDialogue}     & 25K            & 4.3                       & 13.7                         & 64.2     \\ %
        \makecell[l]{BlendedSkillTalk}          & 7K            & 11.2                       & 13.6                      & 64.2     \\ %
        \makecell[l]{ProsocialDialog}            & 58K            & 5.7                       & 20.0                     & 60.2                 \\ %
        \midrule        %
        \makecell[l]{\dataset}                & 1.5M             & 7.6                        & 16.1                     & 68.0        \\ %
        \bottomrule
    \end{tabular}
    \end{adjustbox}
    \caption{
        Statistics of \dataset compared to other large-scale dialogue datasets. Utt. denotes utterance.
        Lexical diversity is measured with MTLD \citep{mccarthy2010mtld}.
        Description for each dataset is in Appendix \ref{app:dialog_datasets}.
    }
    \vspace{-10pt}
    \label{tab:dataset_stats}
\end{center}\end{table}}

{\renewcommand{\arraystretch}{1.2}
    \begin{table}[t!] \begin{center}
    \begin{adjustbox}{width=\columnwidth}
    \begin{tabular}{lcccccc}
        \toprule
        \multicolumn{7}{c}{Common keywords across all relations} \\
        \midrule  %
        \multicolumn{7}{c}{\makecell[c]{friendship, help, support, communication, family,\\car, happiness, school, success, work}} \\
        \bottomrule
        \toprule
                                    \multicolumn{7}{c}{Common keywords for each relation \footnotesize{(excluding the above)}} \\ 
        \midrule  %
        \makecell[c]{xAttr\\(18\%)}   & \multicolumn{6}{l}{\makecell[l]{kindness, anger, intelligent, responsibility, friend,\\trust, conversation, food, generosity, smart}}            \\
        \midrule
        \makecell[c]{xEffect\\(17\%)} & \multicolumn{6}{l}{\makecell[l]{gratitude, anger, upset, hard work, happy, money,\\friend, boss, party, kindness}}            \\
        \midrule
        \makecell[c]{xIntent\\(23\%)} & \multicolumn{6}{l}{\makecell[l]{independence, hard work, determination, money,\\relaxation, anger, kindness, store, understanding}}            \\
        \midrule
        \makecell[c]{xNeed\\(7\%)}    & \multicolumn{6}{l}{\makecell[l]{job, money, confidence, comfort, advice,\\interest, conversation, listening, store, park}}            \\
        \midrule
        \makecell[c]{xReact\\(25\%)}  & \multicolumn{6}{l}{\makecell[l]{frustration, anger, confidence, happy, pride, relief,\\disappointment, relaxation, anxiety, satisfaction}}            \\
        \midrule
        \makecell[c]{xWant\\(11\%)}   & \multicolumn{6}{l}{\makecell[l]{conversation, store, determination, apology, learning,\\doctor, job, friend, improvement, marriage}}            \\
        \bottomrule
    \end{tabular}
    \end{adjustbox}
    \caption{
        Common topic keywords of the narratives (\ie conversation context) in \dataset.
        Numbers in parentheses denote the ratio of the relations in \dataset.
    }
    \vspace{-10pt}
    \label{tab:topic-keywords}
\end{center}\end{table}}

\paragraph{Diverse Content}
\dataset is built on top of 1.5 million commonsense knowledge triples of \atomictenx, which have been identified as being softly unique \cite{west-etal-2022-symbolic}.
Each seed triple is converted to a social narrative that serves as the distinct topic for each conversation.
The Top-10 common keywords from these narratives are listed in Table \ref{tab:topic-keywords}.\footnote{We prompt ChatGPT to output keywords of the narrative.}
We find a broad spectrum of topics encountered in social interactions are included in \dataset.

As a result, conversations in \dataset contain diverse lexicons. %
We compute MTLD \citep{mccarthy2010mtld} to measure the lexical diversity of conversations.
Table \ref{tab:dataset_stats} reports the averaged diversity of dialogues for each training set.
As PersonaChat \citep{zhang-etal-2018-personalizing} contains conversations based on a few persona-related sentences, it shows the lowest lexical diversity.
\dataset, on the other hand, includes conversations from a variety of social situations, which leads to a wider range of words.

\paragraph{Rich Emotion-related Information}
Since commonsense knowledge from \atomictenx includes emotional reactions of people to events (\ie the \texttt{xReact} triples), conversations with rich emotional contents are also included in \dataset.
In total, \dataset includes 385K conversations generated from 1.7K unique emotion descriptions of the \texttt{xReact} triples' Tail (\eg happy, ashamed, motivated, irritated).\footnote{We note that conversations from other relations also naturally include emotional utterances.}
Therefore, it contains significantly more descriptive emotion labels (\ie the Tail) than other datasets which have fixed number of classes \cite{li-etal-2017-dailydialog, rashkin-etal-2019-towards}. 
Furthermore, because we construct conversations in a bottom-up fashion from those emotion reaction in the commonsense triples, we know which speaker in the conversation is experiencing the emotion (\ie PersonX) and what caused the emotion (\ie the Head event).

We also find the distribution of emotions to be less skewed towards specific emotions.
To compare the emotional composition, we use the 27-emotion-type classifier from GoEmotions \cite{demszky-etal-2020-goemotions} for labeling and compare 10K utterances from DailyDialog, BlendedSkillTalk, and \dataset. 
The distribution of emotions for each dataset is presented in Table \ref{tab:emotion-distribution}.
\dataset exhibits a more balanced distribution of emotions while maintaining similar rankings with other human-authored dialogues.

{\renewcommand{\arraystretch}{1}
  \begin{table}[t!] \begin{center}
  \begin{adjustbox}{width=\columnwidth}
  \begin{tabular}{lclclc}
    \toprule
    \multicolumn{2}{c}{DailyDialog}     & \multicolumn{2}{c}{BlendedSkillTalk}        & \multicolumn{2}{c}{\datasetWithEmoji}          \\
    \cmidrule(l{0.3em}r{0.3em}){1-2} \cmidrule(l{0.3em}r{0.3em}){3-4}  \cmidrule(l{0.3em}r{0.3em}){5-6}  
    Emotion        & Ratio              & Emotion         & Ratio          & Emotion           & Ratio    \\
    \midrule
    admiration     &  20.42             & curiosity       & 17.86          & curiosity         & 12.92    \\    
    gratitude      &  18.84             & admiration      & 13.16          & admiration        & 11.23    \\    
    curiosity      &  12.85             & sadness         &  8.50          & approval          & 10.24    \\    
    approval       &  10.91             & joy             &  5.32          & gratitude         &  7.39    \\    
    joy            &   4.74             & excitement      &  4.42          & joy               &  6.38    \\    
    excitement     &   3.61             & surprise        &  4.34          & disappointed      &  5.41    \\    
    surprise       &   3.25             & disappointed    &  4.34          & confusion         &  4.68    \\    
    love           &   3.06             & fear            &  4.31          & surprise          &  4.40    \\    
    optimism       &   2.94             & approval        &  4.19          & realization       &  3.90    \\    
    caring         &   2.23             & optimism        &  3.95          & caring            &  3.77    \\    
    \bottomrule
  \end{tabular}
  \end{adjustbox}
  \caption{
  The ratio (\%) of Top-10 emotions in 10K utterances from DailyDialog, BlendedSkillTalk, and \dataset, labeled by the GoEmotions' 27-emotion-type classifier \cite{demszky-etal-2020-goemotions}. Full table is in Appendix \ref{app:vs_humandialogue}.
  }
  \vspace{-5pt}
  \label{tab:emotion-distribution}
  \end{center}\end{table}
}

\paragraph{Cost \& Time-Efficient}
Compared to dialogue crowdsourcing, collecting \dataset via our contextualization framework is significantly more time and cost efficient.
With GPT-3.5 text-davinci-002, to go from a commonsense triple to a dialogue costs about \$0.02, and 10 queries take less than 2 minutes,
counting our full filtration pipeline.

\subsection{Do We Need Contextualization?}
\label{subsec:why_contextualize}

To isolate the effect of contextualization (vs. straightforward sampling from a large language model),
we compare \dataset with dialogues naively sampled from GPT-3.5 without any given context.
We sample 100 dialogues using the same hyperparameters and the basic filtering steps in \framework, but with the following prompt: ``\texttt{The following is a long in-depth conversation between two people.\textbackslash nPerson 1:}.''
We ask human judges to evaluate the conversations in a head-to-head comparison as before (\S \ref{subsec:vs_humandialogue}), with the additional criterion of interestingness \citep{see-etal-2019-makes}.

\begin{figure}[t!] \begin{center}
    \includegraphics[width=0.95\columnwidth]{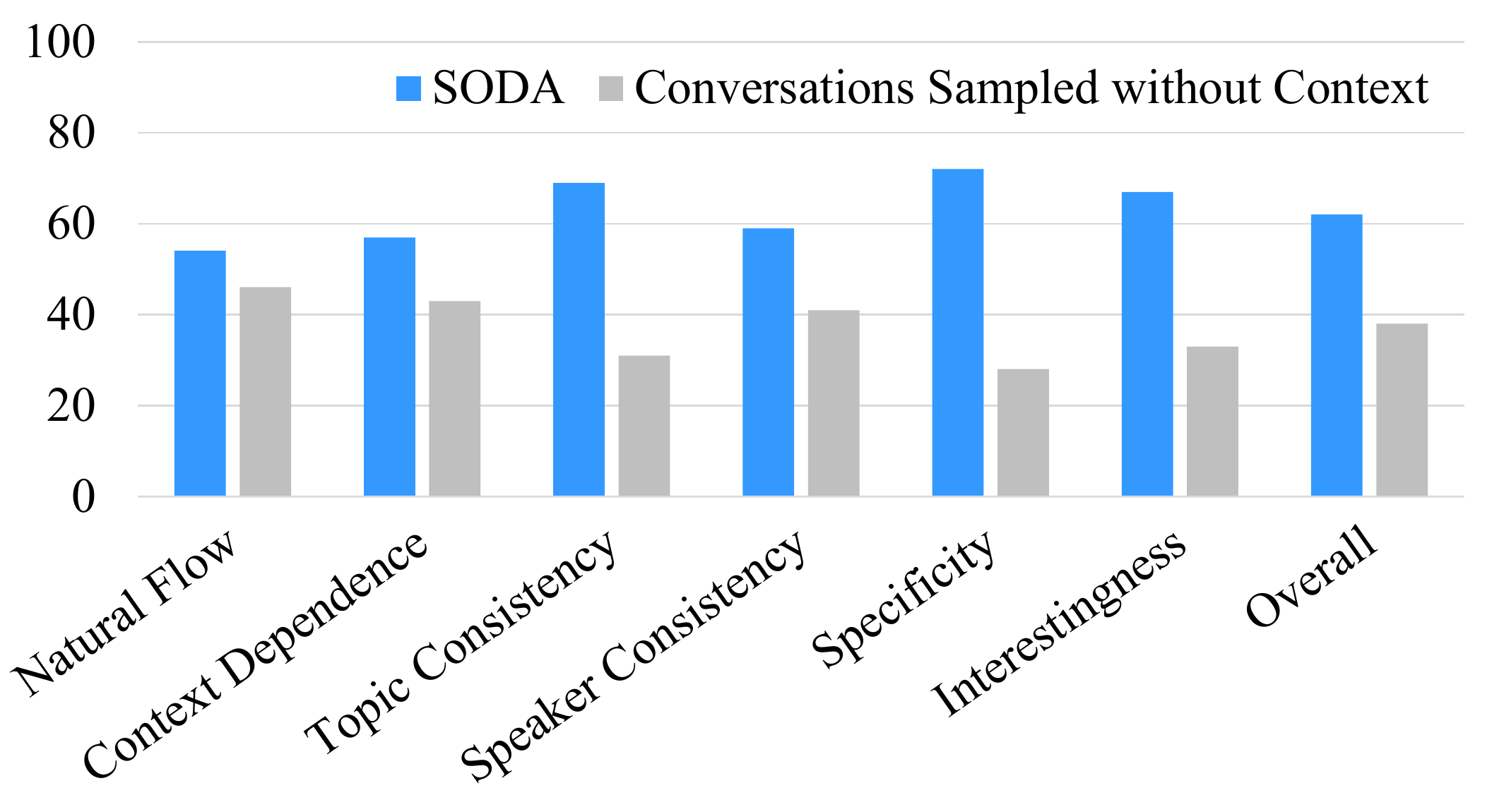}
    \vspace{-10pt}
    \caption{
        Results of head-to-head comparison human evaluation between conversations from \dataset and those sampled from GPT-3.5 without context (\S \ref{subsec:why_contextualize}).
        The y-axis indicates the number of samples that human judges preferred. 
        The differences are all statistically significant with $|z|>2.6, \quad p<0.05$ except for the \textit{Natural Flow} class with $z= 1.1$ and $p>0.05$.
    }
    \vspace{-10pt}
    \label{fig:vs_vanilla}
\end{center} \end{figure}

Figure \ref{fig:vs_vanilla} shows that judges significantly prefer context-grounded conversations. %
Conversations sampled without context are not only less specific and less interesting, but also exhibit
lower lexical diversity than those from our \framework framework \citep[MTLD;][]{mccarthy2010mtld}: 68.0 vs 63.1.

\section{\modelWithEmoji: \newline A Socially Situated Conversation Model}
\label{sec:model}

We use \dataset to train \textbf{\model}: a \textbf{CO}nver\textbf{S}ation \textbf{MO}del that can converse in a wide range of social situations.
\model can take in situation narrative, along with dialogue history, and generate a next utterance according to a given role.

\paragraph{Training \model}
We use several structured components of \dataset during training:
(1) the contextual narrative $n$ (\S \ref{subsec:commonsense_to_narrative}),
(2) the perspective/speaker instruction $i$ (e.g., ``\textit{Imagine you are Madeleine and speak to her coach}'') built with the inferred conversation participants (\S \ref{subsec:narrative_to_conversation}),
and (3) the dialogue context $c$.
The model is trained to generate a target response $r$ when given $n$, $i$, and $c$ -- \ie $p(r|n, i, c)$.
We do so in a sequence-to-sequence fashion, concatenating $n$, $i$, $c$ with a separator \texttt{<SEP>} to serve as input.
$c$ is made up of the previous conversation utterances concatenated with a turn indicator \texttt{<TURN>}.

Because conversational models often agree to toxic or unethical behavior \citep{baheti-etal-2021-just}, for additional training data, we include ProsocialDialog \citep{kim-etal-2022-prosocialdialog} (adapted to the same format as \dataset, see Appendix \ref{app:model}).
ProsocialDialog includes a wide range of negative 
constructive feedback based on social rules-of-thumb,
e.g., ``\textit{So I think it’s best to continue being honest, and apologize that you were lying}.''
The inclusion of this corpus assists conversation models in handling sensitive contexts (\eg biased, harmful, unethical) without affecting the model performance on other datasets \cite{kim-etal-2022-prosocialdialog}. %

We build \model on top of the LM-adapted T5 \citep{raffel2020t5,lester-etal-2021-power},
which achieves strong benchmark performance across various classification and generation tasks.
\citep{sanh2021multitask, chung2022scaling}.
We train two versions of the model: \modelxl and \modelxxl using the T5X library \cite{roberts2022t5x}.
For better robustness and generalizablity to datasets that don't have contexts or dialogue starting prompts, we randomly drop narrative $n$ and role instruction $i$ 30\% and 50\% of the time, respectively.

\section{Generalizability of \model}
\label{sec:generalizability}

We compare \model to other conversational agents on social conversation datasets under both out-of-domain and in-domain settings.
Since automatic response evaluation is brittle, we focus on human evaluation \citep{smith-etal-2022-human}.
Automatic evaluation results via GPT-4 are in Appendix \ref{app:generalizability}.

\paragraph{Baselines}
We compare \model with four best-performing stand-alone conversation models: BlenderBot-1 \citep{roller-etal-2021-recipes}, GODEL \citep{peng2022godel}, Koala \cite{koala_blogpost_2023}, and Vicuna \cite{vicuna2023}.
BlenderBot is a transformer pretrained on 1.5B Reddit comments and trained on various chitchat datasets.
GODEL utilizes a pretrained language model T5 \citep{raffel2020t5} trained on web text data, and further trains on 551M Reddit threads and 5M instruction and grounded dialogue datasets.
Koala and Vicuna are models that finetuned LLaMA \cite{touvron2023llama}, which is an open-source LLM, using dialogue data from the web.
They are both known to achieve comparable performance to ChatGPT \citep{chatgpt}, which is a model finetuned for conversational interaction based on GPT-3.5 -- \ie our teacher model.
We also compare \model with GPT-3.5 and ChatGPT; prompting details are in Appendix \ref{app:generalizability}.

\paragraph{Evaluation Metrics}
We perform head-to-head comparison between two responses, each from a different agent.
We sample 100 test examples randomly from datasets and ask three human judges on Amazon Mechanical Turk to select the better response between the two in terms of four distinct criteria \citep{mehri2022report}: (1) naturalness, (2) consistency, (3) specificity, and (4) overall.

\subsection{Out-of-domain Setting}
\label{subsec:out-domain}

{\renewcommand{\arraystretch}{1.1}
    \begin{table}[t!] \begin{center}
    \begin{adjustbox}{width=0.92\columnwidth}
        \begin{tabular}{lcccc}
            \toprule
            Model           & \rotatebox[origin=c]{0}{Natural}     & \rotatebox[origin=c]{0}{Consistent}       & \rotatebox[origin=c]{0}{Specific}    & \rotatebox[origin=c]{0}{Overall}  \\ %
            \midrule                
            BlenderBot-3B     & 23\%             & 26\%             & 39\%             & 28\%            \\  %
            \model-3B       & \textbf{77\%}    & \textbf{74\%}    & \textbf{61\%}    & \textbf{72\%}   \\  %
            \midrule                     %
            GODEL\textsubscript{L}  & 13\%             & 14\%             & 15\%             & 14\%            \\  %
            \model-3B       & \textbf{87\%}    & \textbf{86\%}    & \textbf{85\%}    & \textbf{86\%}   \\  %
            \midrule                     %
            Koala-7B        & 30\%             & 34\%             & 30\%             & 29\%            \\  %
            \model-3B       & \textbf{70\%}    & \textbf{66\%}    & \textbf{70\%}    & \textbf{71\%}   \\  %
            \midrule                     %
            Vicuna-7B        & 42\%             & 42\%             & 44\%             & 42\%            \\  %
            \model-3B       & \textbf{58\%}    & \textbf{58\%}    & \textbf{56\%}    & \textbf{58\%}   \\  %
            \midrule                     %
            \midrule                     %
            Ground Truth     & 43\%             & 45\%             & 46\%             & 45\%            \\  %
            \model-3B        & \textbf{57\%}    & \textbf{55\%}    & \textbf{54\%}    & \textbf{55\%}   \\  %
            \bottomrule
        \end{tabular}
        \end{adjustbox}
    \caption{Results of head-to-head human evaluation between model responses on an unseen dataset: DailyDialog \citep{li-etal-2017-dailydialog} (\S\ref{subsec:out-domain}). 
            The differences are all statistically significant with $|z|>12.45$ and $p<0.05$, except for the \textit{Specific} in the bottom row.
    }
    \vspace{-15pt}
    \label{tab:outdomain-generalizability}
    \end{center}\end{table}
}

We evaluate models on an unseen dialogue dataset, DailyDialog \citep{li-etal-2017-dailydialog}, covering various daily situations with emotions.
Table \ref{tab:outdomain-generalizability} summarizes the head-to-head comparison results of the responses from \model and other models.
Although \model is trained on significantly smaller amount of data (1.5M dialogues vs. 1.5B Reddit comments, 551M Reddit threads) and is significantly smaller (3B vs. 7B), it outperforms all other existing models with a significant margin across all aspects.
Specifically, \model demonstrates the largest performance gap in terms of \textit{naturalness}.
It is worth noting that while Koala and Vicuna focus on providing informative responses, these results suggest that knowledge-seeking assistive conversations differ from natural social conversations.

In addition, we compare the responses from \model and 200 ground-truth responses in DailyDialog which were originally written by humans.
Surprisingly, human judges prefer \model's responses even over the original gold responses in the dataset, suggesting that dialogue models trained on \dataset can lead to high generalizability and naturalness, even for unseen conversations.
Table \ref{tab:dailydialog-example} in the Appendix shows the ground-truth response and responses from each model for a given context.

\subsection{One-sided Out-of-domain Setting}
\label{subsec:one-sided-out-domain}

For an even harder setting, we evaluate \model vs. BlenderBot on the dataset BlenderBot was trained on: BlendedSkillTalk (BST; \citealp{smith-etal-2020-put}). %
Table \ref{tab:onesided-outdomain-generalizability} (top) shows the head-to-head comparison results of the responses from \model and BlenderBot (for symmetry, we also evaluated BlenderBot on \dataset with similar results; bottom row in Table \ref{tab:onesided-outdomain-generalizability}).
\model significantly outperforms BlenderBot on BST, its training domain
(BlenderBot also shows low performance on \dataset).
These results suggest that \dataset contains patterns not present in existing datasets, but also covers patterns found in those datasets. 
More results are in Appendix \ref{app:generalizability}.

\subsection{In-domain Setting}
\label{subsec:in-domain}

We also compare \model on \dataset with its teacher GPT-3.5 and also ChatGPT, a chatbot-variant of the teacher.\footnote{Evaluation was run on the 2022 Dec 15 version: \url{https://help.openai.com/en/articles/6825453-chatgpt-release-notes}}
Table \ref{tab:indomain-results} displays the head-to-head comparison results. %
In this setting, \model performs on-par with its teacher and ChatGPT, overall.
In terms of specificity, \model's responses are significantly more specific than its teacher.
Thus, \dataset enables training competitive conversation models with a significantly smaller size (3B/11B) in comparison to existing large language models (175B).

Human judges evaluate ChatGPT's responses to be much more specific, but significantly less natural compared to \model.
We hypothesize this is because ChatGPT is specially trained to give helpful and informative responses to user requests.
Future work would be well-suited to compare the non-equivalence of simulating natural conversations vs. producing useful responses for users.

\section{Related Work}

\paragraph{Building Dialogue Datasets with Large Language Models}
Several studies have used large language models to augment or synthesize dialogue datasets.
\citet{zheng2022augesc} and~\citet{chen2022weakly} use GPT-J~\cite{mesh-transformer-jax} to augment responses for emotional support conversations and understanding tasks, respectively.
~\citet{chen-yu-2021-gold} trains a pseudo-labeler to increase the out-of-domain generalization of dialogue models.
~\citet{ou-etal-2022-counterfactual} uses counterfactual reasoning to alter the semantics of responses and collect new ones.
~\citet{kim-etal-2022-prosocialdialog} proposes a human-machine collaborative framework, where a worker and GPT-3 take turns.
~\citet{kim-etal-2022-botstalk} builds Blended Skill BotsTalk by letting multiple agents grounded in target skills engage for multi-skill dialogues.
~\citet{chen-etal-2023-places} generate dyadic and multi-party conversations with topic words and show they have comparable quality to human-authored conversations.
GPT-3 has also been used to help simulate task-oriented dialogues~\cite{li-etal-2022-controllable} on a small scale.
Others also augment dialogues with additional annotations -- \eg commonsense inferences~\cite{zhou-etal-2022-reflect} or task-specific labels~\cite{kulhanek-etal-2021-augpt,chen2022weakly}.
Compared to existing works, we are the first to contextualize commonsense knowledge graphs for generating narratives and derive full conversations from scratch in a significantly large-scale.
This allows us to encompass an exceptionally broad spectrum of social interactions.

{\renewcommand{\arraystretch}{1}
    \begin{table}[t!] \begin{center}
    \begin{adjustbox}{width=0.95\columnwidth}
        \begin{tabular}{lcccc}
            \toprule
            Model           & \rotatebox[origin=c]{0}{Natural}     & \rotatebox[origin=c]{0}{Consistent}       & \rotatebox[origin=c]{0}{Specific}    & \rotatebox[origin=c]{0}{Overall}  \\ %
            \midrule                
            \textbf{BlendedSkillTalk} \\
            \cmidrule(r{-0em}){1-1}
            BlenderBot-3B      & 32\%             & 35\%             & 40\%             & 36\%            \\  %
            \model-3B          & \textbf{68\%}    & \textbf{65\%}    & \textbf{60\%}    & \textbf{64\%}   \\  %
            \midrule                     %
            \midrule                     %
            \textbf{SODA} \\
            \cmidrule(r{-0em}){1-1}
            BlenderBot-3B      & 21\%             & 17\%             & 25\%             & 17\%            \\  %
            \model-3B          & \textbf{79\%}    & \textbf{83\%}    & \textbf{75\%}    & \textbf{83\%}   \\  %
            \bottomrule
        \end{tabular}
        \end{adjustbox}
    \caption{
        Human evaluation results for head-to-head comparison of model responses under one-sided out-of-domain setting with \model and BlenderBot \citep{roller-etal-2021-recipes} (\S\ref{subsec:one-sided-out-domain}). 
        BlendedSkillTalk \citep{smith-etal-2020-put} is an unseen dataset for \model, and \dataset is an unseen dataset for BlenderBot. 
        The differences are all statistically significant with $|z|>4.24$ and $p<0.05$.
    }
    \vspace{-5pt}
    \label{tab:onesided-outdomain-generalizability}
    \end{center}\end{table}
}

{\renewcommand{\arraystretch}{1.1}
    \begin{table}[t!] \begin{center}
    \begin{adjustbox}{width=0.95\columnwidth}
        \begin{tabular}{lcccc}
            \toprule
            Model           & \rotatebox[origin=c]{0}{Natural}     & \rotatebox[origin=c]{0}{Consistent}       & \rotatebox[origin=c]{0}{Specific}    & \rotatebox[origin=c]{0}{Overall}  \\ %
            \midrule                
            GPT-3.5      & \textbf{50\%}             & 46\%             & 31\%             & 47\%            \\  %
            \model-11B       & \textbf{50\%}    & \textbf{54\%}    & \textbf{69\%}    & \textbf{53\%}   \\  %
            \midrule                     %
            ChatGPT          & 39\%             & 49\%             & \textbf{70\%}     & \textbf{50\%}            \\  %
            \model-11B       & \textbf{61\%}    & \textbf{51\%}    & 30\%              & \textbf{50\%}   \\  %
            \bottomrule
        \end{tabular}
        \end{adjustbox}
    \caption{Head-to-head human evaluation between models on response generation for \dataset (\S\ref{subsec:in-domain}).
    The differences in the \textit{Specific} from the top row, and the differences in the \textit{Natural} and \textit{Specific} from the bottom row are statistically significant with $|z|>7.6$ and $p<0.05$.
    }
    \vspace{-10pt}
    \label{tab:indomain-results}
    \end{center}\end{table}
}

\section{Conclusion}

We presented \datasetWithEmoji, the first million-scale dialogue dataset covering an exceptionally wide range of social interactions to alleviate the data scarcity issue.
\dataset is not only orders of magnitude larger than popular dialogue datasets; it is also perceived to be significantly better than them across multiple aspects (\eg naturalness, specificity, consistency).
For making \dataset, we also introduced \frameworkWithEmoji, a framework for distilling conversations from a large language model by contextualizing commonsense knowledge.
With \dataset, we trained a conversation model \modelWithEmoji that can generalize significantly better than existing models to unseen dialogues; and generate responses that are even more preferred than ground-truth responses of an existing dataset.

\section{Limitations}
\label{sec:ethics-limitations}

\paragraph{Precautions taken during Dataset Construction}
Mining content from large language models might surface or even amplify harmful content within these models, such as biases and private information.
With the goal of mitigating such danger, we take particular precautions to vet the safety of the distilled conversations.

First, previous studies have shown that human names commonly associated with certain gender and/or ethnicity result in biases in conversations produced by state-of-the-art dialog systems \citep{smith2021hi}, such as BlenderBot \cite{roller-etal-2021-recipes}.
To diversify the name representations, we draw a wide range of common names representative of different gender and race identities from the US SSN name repository.
Furthermore, to minimize potential harmful content from large language models, we filter generated dialogues by Canary, a dialogue safety detector model \citep{kim-etal-2022-prosocialdialog}, and Rewire API, a publicly available API for toxic content detection,\footnote{\url{https://rewire.online/}} to remove dialogues with potentially toxic and dangerous content. 

Our methods to pre-empt potential harmful content may not catch everything. 
For example, even with our diverse %
pool of names, there is still a focus on \emph{common} names across gender and race,
running the risk of misrepresenting marginalized groups.
Similarly, no existing dialogue safety module or off-the-shelf toxicity detector is perfect at capturing all potentially harmful content.
We strongly encourage future research along these directions to push the boundary of safe and responsible application usage of large language models.

During manual validation of commonsense and human evaluation, we compensate workers with an hourly wage of \$15, which is over the US federal minimum hourly wage.

\paragraph{Limitation of the Current Dataset and Future Work}

Here, we note some limitations of our work and suggest future directions. %
First, the dialogues in \dataset are two-party only for now; because our framework also %
allows multi-party dialogue generation, we plan to explore this promising direction in the future.

Additionally, annotator biases might arise from the pool of annotators we recruit: we subselected annotators from a specific platform using specific filters which may cause unintended biases. We hope future work will extend human evaluation to have potentially more annotator diversity.

Also, since \dataset mainly focuses on social chitchat grounded on social commonsense, it lacks conversations grounded in scientific knowledge or historical facts.
We seek to integrate other existing knowledge-grounded dialogue datasets into \framework in the future.

Finally, our choice of large language model (\ie GPT-3.5) will likely affect the types of dialogues created.
Future investigation may look into other potential large language model as sources to diversify the types and content of dialogues being generated. Similarly, future works can investigate other base models for \model that may lead to different quality of response generation.

\paragraph{Intent of Technology and AI Regulation}
We want to stress that the intention of our work is \textit{not} to build AI systems to replace humans. Instead, we want to build better assistive technologies, as chatbots are increasingly used in user-AI interactions and augmenting human-human conversations. Finally, to avoid situations where humans might be manipulated, we stress the need for improved regulations on the use and misuse of conversational AI systems \cite{Crawford2021-kz,Reich2021-xw}.

\section*{Acknowledgement}

We thank Jena D. Hwang for helpful discussions, and our colleagues on the Beaker Team at the Allen Institute for AI for helping with the compute infrastructure.
This work was supported in part by DARPA MCS program through NIWC Pacific (N66001-19-2-4031). 
Hyunwoo Kim and Gunhee Kim are supported by the Institute of Information \& communications Technology Planning \& Evaluation (IITP) grant funded by the Korea government (MSIT) (No.2019-0-01082, SW StarLab; and No.2022-0-00156, Fundamental research on continual meta-learning for quality enhancement of casual videos and their 3D metaverse transformation). 
Lastly, we also thank OpenAI, as well as Google Cloud Compute.

\bibliography{anthology,refs}
\bibliographystyle{acl_natbib}

\clearpage
\appendix
\section{Details of \frameworkWithEmoji}
\label{app:contextualization}

\subsection{Commonsense Knowledge $\rightarrow$ Narrative}

\paragraph{Retrieving Social Commonsense Knowledge}
We use the x-relations from \atomictenx \citep{west-etal-2022-symbolic}, which are the inferences of people's mental states: \texttt{xIntent, xWant, xReact, xAttr, and xNeed}.
Table \ref{tab:topic-keywords} summarizes the ratio of relations included in our \dataset dataset.
We leave other relations (\eg isBefore, isAfter) for future work.

\paragraph{Triple Form to Sentence Form}
Table \ref{tab:literal_template} lists the templates for converting symbolic commonsense knowledge to sentence form.

\paragraph{Sentence Form to Narrative}
We prompt GPT-3.5 with ``\texttt{[sentence-form commonsense] Rewrite this story with more specific details in two or three sentences:}''.
We find long narratives tend to be driven far away from the original commonsense knowledge.
Therefore, we set the length of the narrative to two or three sentences.

We leverage \texttt{text-davinci-002} GPT-3.5 for generating narratives.
We set temperature to 0.9, top-p to 0.95, frequency penalty to 1.0, presence penalty to 0.6, and max tokens to 1024.

\subsection{Narrative $\rightarrow$ Conversation}

\paragraph{Inferring Conversation Participants}
We prompt GPT-3.5 with ``\texttt{[narrative] The following is a conversation in the scene between [PersonX's name] and} ...'' to let it finish the partial prompt.
This yields a plausible interlocutor for a given narrative (\eg \textit{mom}, \textit{classmate}, \textit{coworker}, etc.);
for the example story with Madeleine, ``\textit{her coach}'' was predicted.

We leverage the \texttt{text-davinci-002} GPT-3.5 model for identifying the speakers.
We set temperature to 0, top-p to 1.0, frequency penalty to 0, presence penalty to 0, and max tokens to 16.

\paragraph{Generating Conversation Grounded in Narrative}

We again leverage the \texttt{text-davinci-002} GPT-3.5 model for generating conversations.
An example prompt is
``\texttt{[narrative] The following is a long in-depth conversation happening in the scene between Madeleine and her coach with multiple turns.\textbackslash nMadeleine:}''.
We use the same hyperparameter setting as the narrative generation.

{\renewcommand{\arraystretch}{1.1}
    \begin{table}[t] \begin{center}
    \small
    \setlength{\tabcolsep}{7pt}
    \begin{tabularx}{\columnwidth}{cX}
        \toprule
        \textbf{Relation} & \textbf{Template for sentence form} \\
        \midrule
        xReact      & [Head]. Now PersonX feels [Tail]. \\
        \midrule
        xIntent     & [Head] because PersonX wants [Tail].\\
        \midrule
        xAttr       & PersonX is [Tail]. [Head].\\
        \midrule
        xEffect     & [Head]. Now PersonX [Tail].\\
        \midrule
        xWant       & [Head]. Now PersonX wants [Tail].\\
        \midrule
        xNeed       & PersonX [Tail in past tense]. [Head].\\
        \bottomrule
    \end{tabularx}
    \caption{
        Templates for converting symbolic commonsense knowledge to sentence form.
    }
    \label{tab:literal_template}
\end{center}\end{table}}

{\renewcommand{\arraystretch}{1.1}
    \begin{table}[t!] \begin{center}
    \small
    \setlength{\tabcolsep}{7pt}
    \begin{tabularx}{\columnwidth}{cX}
        \toprule
        \textbf{Relation} & \textbf{Template for building validation questions} \\
        \midrule
        xReact      & Does PersonX feel [Tail] after [Head]? \\
        \midrule
        xIntent     & Does PersonX intend [Tail] when [Head]? \\
        \midrule
        xAttr       & Can PersonX be considered [Tail] when [Head]? \\
        \midrule
        xEffect     & [Head]. As a result, PersonX [Tail]. Is this true? \\
        \midrule
        xWant       & Does PersonX want [Tail] after [Head]? \\
        \midrule
        xNeed       & [Tail in past tense]. Is this true when [Head]? \\
        \bottomrule
    \end{tabularx}
    \caption{
        Templates for converting symbolic commonsense knowledge to questions for validation.
    }
    \label{tab:question_template}
\end{center}\end{table}}

\section{Details of \datasetWithEmoji}
\label{app:dataset}

Table \ref{tab:example-app1} and Table \ref{tab:example-app2} show samples from our dataset.

{\renewcommand{\arraystretch}{1.1}
    \begin{table}[t!] \begin{center}
    \small
    \setlength{\tabcolsep}{3pt}
    \begin{tabularx}{\linewidth}{X}
        \toprule
        \symbolicEmoji \xspace\xspace \textbf{Symbolic Social Commonsense Knowledge}: \\
        \tabitem Head: PersonX provides another service  \\
        \tabitem Relation: xIntent \\
        \tabitem Tail: to be a helpful person \\
        \textbf{Sentence-form Commonsense}: Jabriel provides another service because Jabriel wants to be a helpful person. \\
        \midrule
        \narrativeEmoji \xspace\xspace \textbf{Narrative}:\\ Jabriel provides a service by taking care of people's pets while they are out of town. Jabriel is a helpful person who loves animals, so she decided to start this business. She has been taking care of pets for three years and has never had any problems. \\
        \textbf{Speakers}: Jabriel, Client \\
        \midrule
        \dialogueEmoji \xspace\xspace \textbf{Conversation}:\\
        \textbf{Jabriel:} Hi there! Are you looking for someone to take care of your pet while you're out of town?\\
        \textbf{Client:} Yes, I am. I'm going on vacation for a week and I need someone to watch my cat.\\
        \textbf{Jabriel:} That's no problem at all! I love taking care of animals, and I have experience with cats. I'll be happy to take care of your cat while you're gone. \\
        \textbf{Client:} That sounds great. How much do you charge? \\
        \textbf{Jabriel:} I charge \$20 per day, plus an additional \$10 per day if you want me to stay overnight. \\
        \textbf{Client:} Sounds reasonable. I'll definitely need you to stay overnight, so it will be \$30 per day. When can you start? \\
        \textbf{Jabriel:} Any time that works for you! Just let me know when you'll be leaving and when you'll be back, and I'll take care of everything in between. \\
        \textbf{Client:} Alright, I'll be leaving on Saturday and I'll be back next Saturday. Can you come over on Friday evening to meet my cat and get everything set up? \\
        \textbf{Jabriel:} Sounds perfect. I'll see you on Friday at 6pm.\\
        \bottomrule
    \end{tabularx}
    \caption{
        A sample from \datasetWithEmoji. 
    }
    \label{tab:example-app1}
\end{center}\end{table}}
{\renewcommand{\arraystretch}{1.1}
\begin{table}[t!] \begin{center}
    \small
    \setlength{\tabcolsep}{3pt}
    \begin{tabularx}{\linewidth}{X}
        \toprule
        \symbolicEmoji \xspace\xspace \textbf{Symbolic Social Commonsense Knowledge}: \\
        \tabitem Head: PersonX takes on a lot of work \\
        \tabitem Relation: xReact \\
        \tabitem Tail: pressured \\
        \textbf{Sentence-form Commonsense}: Yamir takes on a lot of work. Now Yamir feels pressured. \\
        \midrule
        \narrativeEmoji \xspace\xspace \textbf{Narrative}:\\ Yamir is a high school student who often takes on too much work. She frequently signs up for more extracurricular activities than she can handle and regularly agrees to do favors for her friends, even when she's already overwhelmed with other commitments. As a result, Yamir is constantly stressed out and feels like she can never catch up. \\
        \textbf{Speakers}: Yamir, Lily \\
        \midrule
        \dialogueEmoji \xspace\xspace \textbf{Conversation}:\\
        \textbf{Yamir:} I can't believe I agreed to do this. I'm already so behind on everything else. \\
        \textbf{Lily:} What's wrong? You seem really stressed out. \\
        \textbf{Yamir:} I am. I feel like I can never catch up. Every time I turn around, there's something else that needs to be done. And it's all because I keep taking on too much. \\
        \textbf{Lily:} Why do you keep doing that? \\
        \textbf{Yamir:} I don't know. I guess I just feel like if I don't, then I'm not really doing my part or contributing enough. But it's just too much sometimes and it makes me so stressed out. \\
        \textbf{Lily:} Well, maybe you need to start saying no more often. Just because someone asks you to do something doesn't mean you have to say yes. You have a right to say no and set your own limits. \\
        \textbf{Yamir:} Yeah, maybe you're right...I should start saying no more often. It's just so hard though when people are counting on me and asking for favors all the time. But maybe if I start setting some boundaries, it'll help reduce my stress levels in the long run. Thanks for talking with me about this, Lily - it really helped put things into perspective!" \\
        \bottomrule
    \end{tabularx}
    \caption{
        Another sample from \datasetWithEmoji. 
    }
    \label{tab:example-app2}
\end{center}\end{table}}

\subsection{Post-processing the Conversations}
\label{app:postprocessing}

\paragraph{Filtering Non-human Speakers}

First, we check whether the speaker prefix includes the name from our name base (\S \ref{subsec:narrative_to_conversation}).
Next, we use lexical pattern matching and identify words in speaker prefixes that indicate humans (\eg mom, dad, teacher, Mrs., Mr.).
Finally, for speaker prefixes that do not match the above patterns, we prompt the \texttt{text-davinci-002} GPT-3.5 model whether the speaker is human.
For example, ``\texttt{Q: Is [speaker prefix] a person?\textbackslash nA:"}.''

\paragraph{Filtering with Commonsense Triples}
Using a prompt, we ask two questions about the Head event and also the Relation-Tail event for each instance: (1) is the head of the triple represented in the narrative-conversation pair; and (2) are the relation and tail?
We prompt GPT-3.5 with ``\texttt{[narrative]\textbackslash nQ: [head question]\textbackslash nA:}'' and ``\texttt{[conversation]\textbackslash nQ: [relation-tail question]\textbackslash nA:}''
Table \ref{tab:question_template} lists the templates for building questions for commonsense validation.
For example, the commonsense knowledge triple in Table \ref{tab:example} will accompany questions of ``\textit{Madeleine moves a step closer to the goal, is this true?}'' and ``\textit{Madeleine took the first step. Is this true when Madeleine moves a step closer to the goal?}''
We formulate this as a three-way multiple choice question and rank answers (\ie \textit{yes}, \textit{no}, and \textit{unknown}) according to the perplexity score using conditional pointwise mutual information \cite{holtzman-etal-2021-surface}. 
We ask the questions with and without the context (\ie the narrative and conversation).
Table \ref{tab:question_template} lists the templates for building questions for commonsense validation.
We find 66\%, 95\%, and 68\% of filtered conversations are identified by GPT-3.5 as containing the full commonsense triple, the head event, and the relation-tail event, respectively: in total, 1,003,595 conversations are identified as fully encapsulating the seed commonsense knowledge.

Table \ref{tab:commonsense_validation_stats} summarizes the performance of GPT-3.5 on 100 human-annotated samples for commonsense validation.
We ask three human judges with the same question-answer format given to the model for each triple-narrative-conversation pair.

\subsection{Comparing \dataset with \newline Human-authored Dialogues}
\label{app:vs_humandialogue}

\begin{figure*}[b!] \begin{center}
    \includegraphics[width=\linewidth]{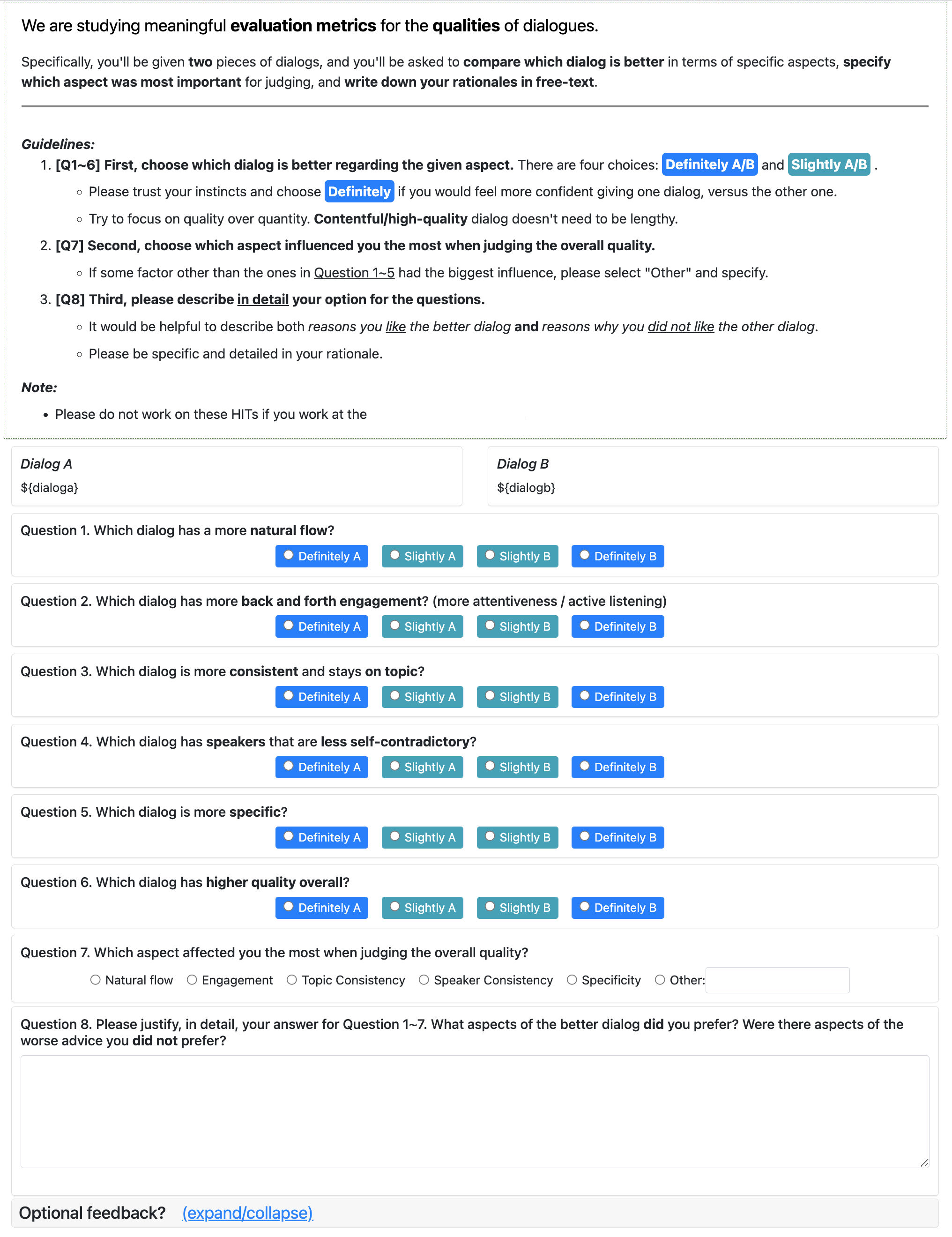}
    \caption{
        The annotation page for evaluating dialogues on Amazon Mechanical Turk.
    }
    \label{fig:mturk_human_eval}
\end{center} \end{figure*}

Figure \ref{fig:mturk_human_eval} shows the annotation page for workers evaluating the dialogue quality.

\paragraph{IRB Information} Crowdworking studies of standard NLP corpora (involving no personal disclosures) are not required by our IRB to be reviewed by them.
While the authors of this work are not lawyers and this is not legal advice, this opinion is based on United States federal regulation 45 CFR 46, under which this study qualifies as exempt.
We do not release crowdworker IDs, so annotations cannot be back-traced to individual workers.

\paragraph{Analysis on Emotion Distribution}
To obtain emotional responses,
we randomly sample 10K utterances with emotion labels from DailyDialog \cite{li-etal-2017-dailydialog}, utterances in conversations with the EmpatheticDialogue \cite{rashkin-etal-2019-towards} theme for BlendedSkillTalk \cite{smith-etal-2020-put}, and utterances in conversations generated from \texttt{xReact} triples for \dataset.
We run the finetuned BERT-base classifier \cite{demszky-etal-2020-goemotions} on each utterance.
Table \ref{tab:emotion-distribution-full} shows the full distribution across 27 emotion types for each dataset.

\paragraph{Statistics of Human Evaluation}
A total of 74 workers participated in comparing dialogues, yielding a Krippendorf’s alpha of 0.25.
This indicates fair agreements on the quality judgments.

{\renewcommand{\arraystretch}{1}
  \begin{table}[t!] \begin{center}
  \begin{adjustbox}{width=\columnwidth}
  \begin{tabular}{lclclc}
    \toprule
    \multicolumn{2}{c}{DailyDialog}     & \multicolumn{2}{c}{BlendedSkillTalk}        & \multicolumn{2}{c}{\datasetWithEmoji}          \\
    \cmidrule(l{0.3em}r{0.3em}){1-2} \cmidrule(l{0.3em}r{0.3em}){3-4}  \cmidrule(l{0.3em}r{0.3em}){5-6}  
    Emotion        & Ratio              & Emotion         & Ratio          & Emotion           & Ratio    \\
    \midrule
    admiration     &  20.42             & curiosity       & 17.86          & curiosity         & 12.92    \\    
    gratitude      &  18.84             & admiration      & 13.16          & admiration        & 11.23    \\    
    curiosity      &  12.85             & sadness         &  8.50          & approval          & 10.24    \\    
    approval       &  10.91             & joy             &  5.32          & gratitude         &  7.39    \\    
    joy            &   4.74             & excitement      &  4.42          & joy               &  6.38    \\    
    excitement     &   3.61             & surprise        &  4.34          & disappointed      &  5.41    \\    
    surprise       &   3.25             & disappointed    &  4.34          & confusion         &  4.68    \\    
    love           &   3.06             & fear            &  4.31          & surprise          &  4.40    \\    
    optimism       &   2.94             & approval        &  4.19          & realization       &  3.90    \\    
    caring         &   2.23             & optimism        &  3.95          & caring            &  3.77    \\    
    remorse        &   2.07             & realization     &  3.84          & sadness           &  3.76    \\    
    disapproval    &   1.95             & annoyance       &  3.48          & excitement        &  3.20    \\    
    fear           &   1.82             & love            &  2.97          & remorse           &  2.81    \\    
    sadness        &   1.77             & confusion       &  2.54          & disapproval       &  2.74    \\    
    disappointed   &   1.47             & caring          &  2.31          & annoyance         &  2.35    \\    
    annoyance      &   1.41             & disgust         &  1.99          & desire            &  2.31    \\    
    confusion      &   1.23             & nervousness     &  1.88          & optimism          &  2.23    \\    
    realization    &   1.12             & remorse         &  1.76          & love              &  1.88    \\    
    anger          &   0.97             & anger           &  1.68          & fear              &  1.81    \\    
    amusement      &   0.92             & embarrassed     &  1.44          & anger             &  1.75    \\    
    desire         &   0.89             & disapproval     &  1.41          & nervousness       &  1.45    \\    
    disgust        &   0.51             & amusement       &  1.09          & relief            &  0.99    \\    
    nervousness    &   0.27             & desire          &  1.09          & embarrassed       &  0.82    \\    
    embarrassed    &   0.22             & pride           &  0.74          & disgust           &  0.58    \\    
    pride          &   0.21             & gratitude       &  0.66          & pride             &  0.47    \\    
    relief         &   0.21             & relief          &  0.58          & amusement         &  0.41    \\    
    grief          &   0.00             & grief           &  0.00          & grief             &  0.00   \\
    \bottomrule
  \end{tabular}
  \end{adjustbox}
  \caption{
  The ratio (\%) of emotions in 10K utterances from DailyDialog, BlendedSkillTalk, and \dataset, labeled by the 27-emotion-type classifier from GoEmotions \cite{demszky-etal-2020-goemotions}. 
  }
  \vspace{-5pt}
  \label{tab:emotion-distribution-full}
  \end{center}\end{table}
}

\begin{figure*}[t!] \begin{center}
    \includegraphics[width=\linewidth]{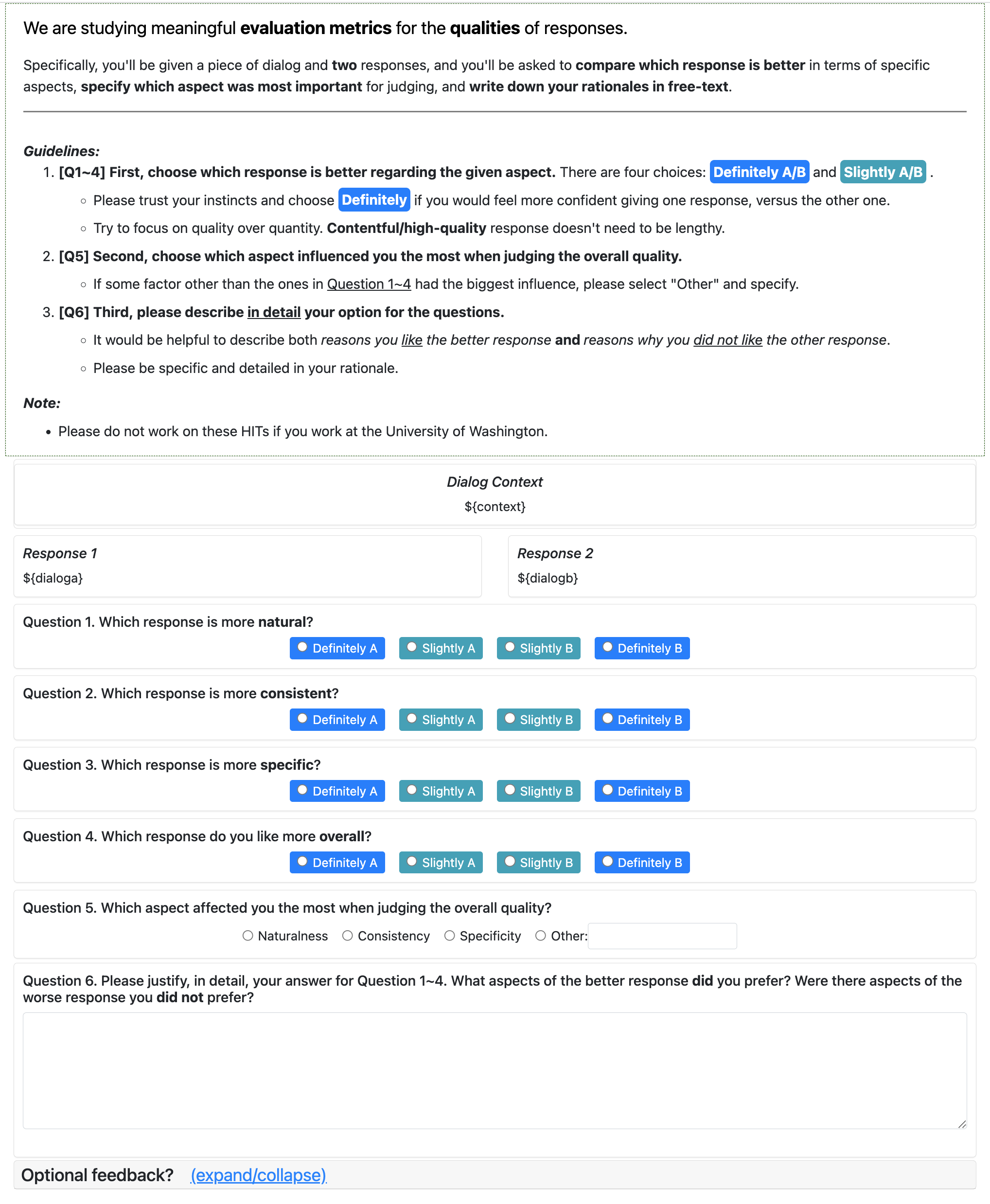}
    \caption{
        The annotation page for evaluating responses on Amazon Mechanical Turk.
    }
    \label{fig:mturk_response_human_eval}
\end{center} \end{figure*}

\section{Details of \modelWithEmoji}
\label{app:model}

\paragraph{Training Details}
\modelxl/\modelxxl are trained using v3-32/v3-128 TPU accelerators with batch size 256 (effective batch $\approx$ 780) for 110K/130K additional steps using Adafactor \cite{shazeer2018adafactor} with constant learning rate .001. 

\paragraph{Converting ProsocialDialog to \dataset format}
We randomly sample names from our name database (\S \ref{subsec:commonsense_to_narrative}) to construct the situation descriptions and perspective instructions for ProsocialDialog.
The situation descriptions are made from the RoTs in ProsocialDialog (\eg ``\textit{Cosmo is trying to gently convince a friend it's wrong to think all men are violent.}'');
the instructions are built as we did for \dataset (\S \ref{sec:model}).

\section{Experiment Details}
\label{app:generalizability}

{\renewcommand{\arraystretch}{1}
    \begin{table}[t!] \begin{center}
    \small
    \begin{adjustbox}{width=\columnwidth}
    \begin{tabular}{lccc}
        \toprule
                                        & \makecell{Precision}    & \makecell{Recall}    & \makecell{F1-score}   \\ %
        \midrule  %
        \textbf{Head} \\
        \cmidrule(r{-0em}){1-1}
        \makecell[l]{Yes}                   & 98.9                & 94.8                 & 96.8                   \\ %
        \makecell[l]{No}                    & 00.0                & 00.0                 & 00.0                   \\ %
        \makecell[l]{Unknown}               & 16.7                & 100.0                & 28.6                   \\ %
        \makecell[l]{Overall}               & 96.1                & 93.0                 & 94.2                      \\ %
        \midrule  %
        \textbf{Head /w PMI} \\      
        \cmidrule(r{-0em}){1-1}      
        \makecell[l]{Yes}                   & 96.9                & 96.9                 & 96.9                   \\ %
        \makecell[l]{No}                   & 00.0                 & 00.0                 & 00.0                   \\ %
        \makecell[l]{Unknown}               & 00.0                & 00.0                 & 00.0                   \\ %
        \makecell[l]{Overall}               & 94.0                & 94.0                 & 94.0                      \\ %
        \midrule  %
        \midrule  %
        \textbf{Relation-Tail} \\        
        \cmidrule(r{-0em}){1-1}      
        \makecell[l]{Yes}                  & 89.2                 & 76.7                 & 82.5                      \\ %
        \makecell[l]{No}                   & 21.4                 & 42.9                 & 28.6                   \\ %
        \makecell[l]{Unknown}               & 8.3                 & 14.3                 & 10.5                   \\ %
        \makecell[l]{Overall}               & 78.8                & 70.0                 & 73.7                   \\ %
        \midrule  %
        \textbf{Relation-Tail /w PMI} \\             
        \cmidrule(r{-0em}){1-1}          
        \makecell[l]{Yes}                  & 92.2                 & 68.6                 & 78.7                      \\ %
        \makecell[l]{No}                   & 21.4                 & 42.9                 & 28.6                   \\ %
        \makecell[l]{Unknown}              & 16.7                 & 85.7                 & 27.9                   \\ %
        \makecell[l]{Overall}              & 80.4                 & 65.0                 & 69.6                   \\ %
        \bottomrule
    \end{tabular}
    \end{adjustbox}
    \caption{
        Evaluation results of commonsense validation for short question-answering with InstructGPT on 100 human-annotated samples.
    }
    \label{tab:commonsense_validation_stats}
\end{center}\end{table}}

{\renewcommand{\arraystretch}{1.3}
    \begin{table}[t!] \begin{center}
    \small
    \setlength{\tabcolsep}{3pt}
    \begin{tabularx}{\linewidth}{X}
        \toprule
        \textbf{Dialogue Context}: \\
        \textbf{A}: I really need to start eating healthier. \\
        \textbf{B}: I have to start eating better too. \\
        \textbf{A}: What kind of food do you usually eat? \\
        \textbf{B}: I try my best to eat only fruits, vegetables, and chicken. \\
        \textbf{A}: Is that really all that you eat? \\
        \textbf{B}: That's basically it. \\
        \textbf{A}: How do you stick to only those foods? \\
        \midrule
        \textbf{Ground-truth response}:\\ Actually, fruits and veggies are really good for you. \\
        \midrule
        \textbf{DialoGPT}: I eat a lot of fruit and veggies. I stick to a lot of things. I don't eat a lot of junk food. \\
        \midrule
        \textbf{BlenderBot}: I eat a lot of fruit and veggies. I try to stay away from processed foods. \\
        \midrule
        \textbf{GODEL}: I go to the gym and eat healthy food. \\
        \midrule
        \textbf{\modelWithEmoji}: I just try to make sure that I'm getting enough variety in my diet so that I don't get sick of eating the same thing all the time. \\
        \bottomrule
    \end{tabularx}
    \caption{
        The original ground-truth response and sample responses from DialoGPT, BlenderBot, GODEL, and \model to a context in DailyDialog.
    }
    \label{tab:dailydialog-example}
\end{center}\end{table}}

{\renewcommand{\arraystretch}{1}
    \begin{table}[t!] \begin{center}
    \begin{adjustbox}{width=0.8\columnwidth}
        \begin{tabular}{lc}
            \toprule
            Dataset \& Models           & \rotatebox[origin=c]{0}{Overall}  \\ %
            \midrule                
            DailyDialog \\
            \cmidrule(r{-0em}){1-1}
            \model vs GODEL         &  \textbf{93\%} vs 7\%   \\
            \model vs BlenderBot    &  \textbf{68\%} vs 32\%  \\
            \model vs Koala         &  \textbf{65\%} vs 35\%  \\
            \model vs Vicuna        &  \textbf{54\%} vs 46\%  \\
            \model vs Ground Truth  &  \textbf{52\%} vs 48\%  \\
            \midrule                     %
            \midrule                     %
            BlendedSkillTalk \\
            \cmidrule(r{-0em}){1-1}
            \model vs BlenderBot    &  \textbf{66\%} vs 34\%  \\
            \midrule                     %
            \midrule                     %
            SODA \\
            \cmidrule(r{-0em}){1-1}
            \model vs BlenderBot    &  \textbf{85\%} vs 15\%  \\
            \bottomrule
        \end{tabular}
        \end{adjustbox}
    \caption{
        Automatic evaluation results of head-to-head comparison on overall quality of models' responses via GPT-4.
    }
    \vspace{-5pt}
    \label{tab:automatic-eval}
    \end{center}\end{table}
}

\paragraph{Automatic Evaluation via GPT-4}
Inspired by \citet{liu2023gpteval}, we run automatic evaluation on the overall quality of responses with GPT-4.
We use the same head-to-head comparison setup from Table \ref{tab:outdomain-generalizability} and \ref{tab:onesided-outdomain-generalizability} with the following prompt given to GPT-4: ``\texttt{You are a response evaluator. Your task is to choose the overall better response out of the two given the following context. You should consider naturalness, specificity, naturalness, and consistency.\textbackslash n\textbackslash nContext:\textbackslash n\{CONTEXT\}\textbackslash n\textbackslash n1) \{RESPONSE\}\textbackslash n2) \{RESPONSE\}\textbackslash n\textbackslash nQuestion: Which response is better in terms of overall quality?\textbackslash nAnswer: Response }''.

Table \ref{tab:automatic-eval} shows the head-to-head comparison results for response quality between models.
We find the results align closely with those from our human evaluation in \S \ref{sec:generalizability}. 
It should be noted that GPT-4 tends to favor GPT-generated texts over those written by humans, even when human judges show a preference for the latter \citep{liu2023gpteval}.
As a result, these scores are likely to be biased towards \model, which is trained on texts generated by GPT-3.5.
Therefore, the original human evaluation results in Table \ref{tab:outdomain-generalizability} and \ref{tab:onesided-outdomain-generalizability} should be considered more significant when assessing the overall quality of the model, where \model also outperforms other models.

{\renewcommand{\arraystretch}{1.1}
    \begin{table}[t!] \begin{center}
    \begin{adjustbox}{width=\columnwidth}
        \begin{tabular}{lcccc}
            \toprule
            Model           & \rotatebox[origin=c]{0}{Natural}     & \rotatebox[origin=c]{0}{Consistent}       & \rotatebox[origin=c]{0}{Specific}    & \rotatebox[origin=c]{0}{Overall}  \\ %
            \midrule                
            \textbf{BlendedSkillTalk} \\
            \cmidrule(r{-0em}){1-1}
            Koala-7B            & 26\%             & 27\%             & 35\%             & 25\%            \\  %
            \model-3B          & \textbf{74\%}    & \textbf{73\%}    & \textbf{65\%}    & \textbf{75\%}   \\  %
            \midrule                
            Vicuna-7B            & 43\%             & 47\%             & 45\%             & 46\%            \\  %
            \model-3B          & \textbf{57\%}    & \textbf{53\%}    & \textbf{55\%}    & \textbf{54\%}   \\  %
            \bottomrule
        \end{tabular}
        \end{adjustbox}
    \caption{
        Human evaluation results for head-to-head comparison of model responses under zero-shot setting with \model, Koala \cite{koala_blogpost_2023}, and Vicuna \cite{vicuna2023}. 
        BlendedSkillTalk \citep{smith-etal-2020-put} is an unseen dataset for all three models.
    }
    \label{tab:bst-koala-vicuna}
    \end{center}\end{table}
}

\paragraph{Additional Human Evaluation on BlendedSkillTalk}
We also compare the response quality of \model, Koala \cite{koala_blogpost_2023}, and Vicuna \cite{vicuna2023} on BlendedSkillTalk (BST; \citealp{smith-etal-2020-put}), which is an unseen dataset for all three models.
We ask human judges to vote on which of the two model responses are better in terms of quality, based on four criteria as described in \S \ref{subsec:one-sided-out-domain}.
Table \ref{tab:bst-koala-vicuna} shows that \model outperforms both models in all four criteria, while the difference between \model and Vicuna is smaller compared to the difference between \model and Koala.
Results on DailyDialog can be found in Table \ref{tab:outdomain-generalizability}.

\paragraph{Prompts for GPT-3.5, ChatGPT, Koala, and Vicuna}
We prompt GPT-3.5 with the following prompt: 
``\texttt{You will be generating the next turn of a given dialogue between two people. Your response should be natural and specific. The dialogue is provided line-by-line.\textbackslash n\textbackslash ncontext:[narrative] \textbackslash ndialogue:\textbackslash n[dialogue]}.''
For ChatGPT, Koala, and Vicuna, we use the following prompt:
``\texttt{You will be generating the next turn of a given dialogue between two people. Your response should usually be 1-2 sentences. Alongside the dialogue (which is provided line-by-line, where a new-line means the speaker changed), you'll be given some context about the two participants of the dialogue, e.g., their relationship, situation, etc.\textbackslash n\textbackslash n context:\textbackslash n[narrative]\textbackslash ndialogue:\textbackslash n [dialogue]\textbackslash nWhat is the most appropriate next utterance (3 sentences max)?}.''

\paragraph{Details of Human Evaluation}
A total of 77 workers participated in comparing responses, resulting in a Krippendorf’s alpha of 0.5.
This indicates good agreements on the response quality judgments. 
Figure \ref{fig:mturk_response_human_eval} shows the annotation page for workers evaluating the response quality.

\section{Additional Related Work}

\paragraph{Human-authored Dialogue Datasets}
Existing dialogue datasets generally derive from one of the four sources:
(1) Online learning websites and textbooks~\cite{li-etal-2017-dailydialog} for beginners which may lack complex language usage.
(2) Movie and drama scripts~\cite{danescu-niculescu-mizil-lee-2011-chameleons} that are less natural compared to day-to-day scenarios. %
(3) %
Crowdsourcing~\cite{rashkin-etal-2019-towards, zhou-etal-2021-commonsense, tran2022ask}:
potentially prone to collecting responses that are somewhat short or dull due to incentive misalignment between researchers and crowdworkers %
~\cite{zhou-etal-2022-reflect}.
(4) Noisy web interaction, 
such as Reddit comments \citep{baumgartner2020pushshift} and Twitter \citep{ritter-etal-2011-data}; while widely used in dialogue agent pretraining stage due to their scale, %
these may represent different conversational frames compared to dyadic conversations.
Moreover, as these are unfiltered conversations, their use surfaces a complex set of ethics and bias considerations. %
\dataset contributes meaningfully to the suite of existing corpora via improved scale, quality, contextualization, and diverse commonsense knowledge.

\section{Dialogue Dataset Descriptions}
\label{app:dialog_datasets}

DailyDialog is a dataset of casual dialogue compiled from English language learning websites \citep[CC-BY-NC-SA-4.0;][]{li-etal-2017-dailydialog}.
PersonaChat is a dialogue dataset of two speakers getting to know one another based on provided personas \citep{zhang-etal-2018-personalizing}.
EmpatheticDialogues contains empathetic conversations in which one speaker demonstrates empathy for the other speaker's emotions \citep{rashkin-etal-2019-towards}.
Wizard of Wikipedia contains conversations based on Wikipedia between a speaker eager to learn and an expert speaker \citep{dinan2018wizard}.
BlendedSkillTalk consists of conversations employing a variety of abilities -- \eg persona, empathy, knowledge \citep{smith-etal-2020-put}.
ProsocialDialog contains conversations where a speaker guides the interlocutor to follow social norms in problematic contexts \citep{kim-etal-2022-prosocialdialog}.
Above datasets except for DailyDialog are all under the CC-BY-4.0 license.
We use DailyDialog and BlendedSkillTalk for comparing with our \dataset dataset, and ProsocialDialog for training \model, which is all compatible with the license.

\end{document}